\documentclass[lettersize,journal]{IEEEtran}

\usepackage{amsmath,amsfonts,amssymb}
\usepackage{bm}
\usepackage{array}
\usepackage{textcomp}
\usepackage{stfloats}
\usepackage{url}
\usepackage{verbatim}
\usepackage{graphicx}
\usepackage{cite}
\usepackage{multirow}
\usepackage{makecell}
\usepackage{booktabs}
\usepackage[table]{xcolor}
\usepackage[caption=false,font=normalsize,labelfont=sf,textfont=sf]{subfig}
\usepackage{capt-of} 

\definecolor{ourcfg}{RGB}{222,235,247}

\begin{document}

\title{See the Change, Keep the Flow: Unsupervised Action Segmentation via Spectral-Temporal Representation Learning}

\author{Yun Li,~\IEEEmembership{Student Member,~IEEE,}
Jun Xiao,~\IEEEmembership{Member,~IEEE,}
Cong Zhang,~\IEEEmembership{Member,~IEEE,} \\
Kin-Man Lam,~\IEEEmembership{Senior Member,~IEEE}

\thanks{Y. Li, J. Xiao and K-M. Lam are with the Department of Electrical and Electronic Engineering, The Hong Kong Polytechnic University, Hong Kong
(e-mail: yun-eie.li@connect.polyu.hk, jun.xiao@connect.polyu.hk,enkmlam@polyu.edu.hk)}
\thanks{C. Zhang is with the School of Control Science and Engineering,
Shandong University (e-mail: congzhang@sdu.edu.cn)}
}

\markboth{Journal of \LaTeX\ Class Files,~Vol.~14, No.~8, August~2021}%
{Shell \MakeLowercase{\textit{et al.}}: A Sample Article Using IEEEtran.cls for IEEE Journals}


\maketitle

\begin{abstract}
Unsupervised action segmentation aims to discover latent action categories and their temporal organization without action annotations. Optimal transport-based methods provide structured frame-to-action assignments, however, their pseudo-label quality is fundamentally conditioned on the representation space used to construct the transport cost. We argue that reliable OT pseudo-labeling requires a representation geometry that is simultaneously sensitive to discriminative action changes and coherent along local temporal progressions. Based on this insight, we propose SpecT-OT, a spectral-temporal representation learning framework built upon an unbalanced optimal transport pseudo-labeling concept. SpecT-OT introduces a Spectral Reparameterization Projector (SRP), which parameterizes projector weights with fixed Fourier bases and learnable coefficients to improve the modeling of rapidly varying discriminative features, and Temporal Affinity Regularization (TAR), which imposes distance-aware, label-free constraints on pairwise frame affinities to stabilize local temporal structure. 
The two components jointly produce more discriminative and temporally stable transport costs, yielding more reliable pseudo-labels for iterative representation learning. Experiments on four benchmarks demonstrate strong performance compared with state-of-the-art methods. SpecT-OT achieves the best results on 13 of 15 metrics, including 4.1-point MoF and 7.4-point F1 gains over the baseline on Breakfast and Desktop Assembly, respectively.
\end{abstract}

\section{Introduction}
Temporal action segmentation aims to partition a long, untrimmed
video into temporally contiguous action segments and assign an action
or sub-activity label to each frame
\cite{chao2018rethinking, ju2024deep, shou2017cdc}. Unlike clip-level action
recognition, it requires both discovering frame-wise action categories
and locating their temporal transitions. Fully supervised methods rely
on dense frame-level annotations, while weakly supervised approaches
still require transcripts, ordered action lists, or timestamps.
Because such annotations are costly for long videos, unsupervised
action segmentation seeks to discover latent action categories and
their temporal organization without action labels during training.

\begin{figure}[t]
    \centering
    \includegraphics[width=1\linewidth]
    {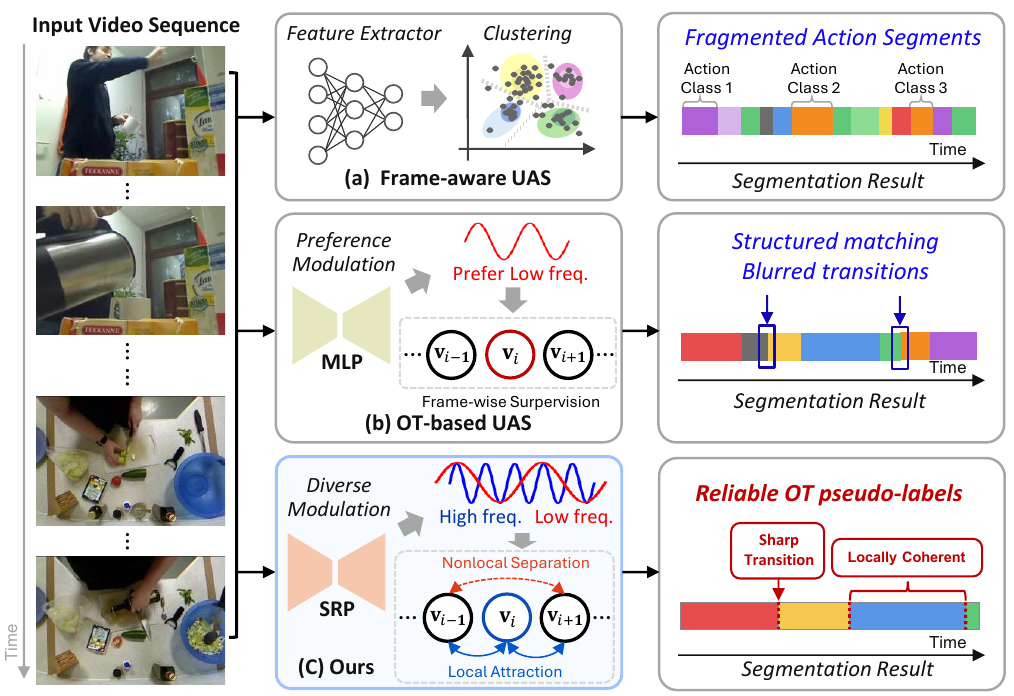}
    \vspace{-0.66cm}
    \caption{Comparison of unsupervised action segmentation paradigms.
    (a) Frame-aware methods often produce fragmented action segments.
    (b) OT-based methods provide structured frame-to-action matching but may blur action boundaries.
    (c) SpecT-OT combines SRP and TAR to preserve discriminative action
    transitions and local temporal coherence, yielding more reliable OT
    pseudo-labels.}
    \label{Intro_feagure}
    \vspace{-0.5cm}
\end{figure}

A common solution is to formulate unsupervised action segmentation as
a task of joint representation learning and latent action assignment. Given
frame features
$\mathbf{X}=\{\mathbf{x}_{t}\}_{t=1}^{L}$, a projector
$f_{\theta}$ produces frame embeddings
$\mathbf{z}_{t}=f_{\theta}(\mathbf{x}_{t})$, which are matched with
$K$ learnable action prototypes
$\mathbf{U}=\{\mathbf{u}_{k}\}_{k=1}^{K}$. Optimal transport (OT) can
then infer the soft frame-to-action assignment as, 
\begin{equation}
    \begin{aligned}
        \mathbf{T}^{*}
        &=
        \arg\min_{\mathbf{T}\geq 0}
        \left\langle
        \mathbf{C}_{\theta,\mathbf{A}},\mathbf{T}
        \right\rangle
        +
        \Omega_{\mathrm{OT}}(\mathbf{T}),
        \\
        C_{tk}
        &=
        1-\operatorname{sim}
        \left(
        f_{\theta}(\mathbf{x}_{t}),
        \mathbf{u}_{k}
        \right),
    \end{aligned}
    \label{eq:intro_ot}
\end{equation}
where $\mathbf{C}_{\theta,\mathbf{A}}$ denotes the frame-action
transport cost, $\Omega_{\mathrm{OT}}(\mathbf{T})$ encodes the
marginal and structural constraints imposed on the transport plan,
and the optimized plan $\mathbf{T}^{*}$ is used as pseudo-label
supervision for representation learning. Methods such as UDE
\cite{swetha2021unsupervised} learn discriminative embeddings for
latent action discovery, while TOT \cite{kumar2022unsupervised}
introduces regularized optimal transport for online pseudo-label
generation.

Standard balanced OT assumes a fixed action-side marginal
distribution, encouraging different action prototypes to receive
similar amounts of transport mass. This assumption is unsuitable for
untrimmed videos, where action durations vary substantially and some
actions may be absent from individual sequences. ASOT
\cite{xu2024temporally} addresses this issue through temporally
consistent unbalanced OT, relaxing the action-side marginal constraint
to accommodate unequal action durations and missing actions.
Nevertheless, relaxing the marginal constraints does not inherently
ensure reliable pseudo-labels. The transport plan is jointly
determined by the OT constraints and the learned cost matrix
$\mathbf{C}_{\theta,\mathbf{A}}$. Even a properly regularized OT
solver may therefore produce unreliable assignments when the
projected feature space yields ambiguous frame--action similarities
or locally unstable frame representations. Since the assignments are
subsequently reused as training targets, early representation errors
may be reinforced through iterative self-training.

A suitable representation space for action segmentation must satisfy
two complementary requirements. First, it should remain sensitive to
feature variations that are informative for distinguishing action
changes. Otherwise, different actions may become difficult to
separate around temporal transitions. Second, it should preserve
local coherence across neighboring frames, preventing minor feature
fluctuations from producing fragmented assignments. In other words,
the model must be able to \emph{see the change} between actions while
\emph{keeping the flow} within locally continuous activity
progressions. Recent OT-based methods mainly focus on improving the marginal or
structural constraints imposed on the transport plan, while paying
less attention to the representation space from which the transport
cost is constructed. Conventional MLP projectors may exhibit a
low-frequency learning preference, limiting their capacity to model
rapidly varying discriminative features. Meanwhile, temporal
constraints applied only during OT inference may not be sufficiently
internalized into the learned frame embeddings, leading to locally
inconsistent similarities and frequent pseudo-label switching. These
two limitations respectively weaken the model's ability to recognize
action changes and maintain temporal continuity.

Based on these observations, we formulate the remaining challenge as
learning a transition-discriminative and locally coherent
representation space underlying OT pseudo-label generation. To this
end, we propose SpecT-OT, a spectral-temporal representation learning
framework built upon unbalanced optimal transport. SpecT-OT introduces
a Spectral Reparameterization Projector (SRP) and Temporal Affinity
Regularization (TAR) to improve the representation space used for
frame--action matching. Specifically, SRP parameterizes the projector weights using fixed
Fourier bases and learnable coefficients. This frequency-diverse
parameterization helps mitigate the low-frequency learning preference
of vanilla MLPs and improves their capacity to model rapidly varying
discriminative features, allowing the projector to better capture
action changes. By providing access to both low- and high-frequency 
feature components, SRP produces more discriminative frame embeddings 
and facilitates more reliable pseudo-label generation. 
Complementarily, TAR directly regularizes pairwise
frame affinities using hard local temporal masks and distance-aware
soft weights. It encourages locally coherent frame representations
and reduces unstable pseudo-label switching, thereby maintaining the
temporal flow of predicted action segments. Together, SRP and TAR
improve frame--prototype discrimination and representation-level
temporal stability, leading to more reliable pseudo-labels for
unsupervised action segmentation.

The contributions of this paper are summarized as follows:
\begin{itemize}
    \item We propose SpecT-OT, a spectral-temporal representation
    learning framework built upon unbalanced OT. SpecT-OT improves
    pseudo-label reliability by jointly enhancing sensitivity to
    action changes and local temporal coherence.

    \item We introduce a Spectral Reparameterization Projector (SRP)
    that represents projection weights as learnable combinations of Fourier bases and
    parameterized by learnable coefficients. This helps mitigate the low-frequency
    preference of the vanilla projector and improve the modeling of rapid
    discriminative feature variations.

    \item We develop a label-free Temporal Affinity Regularization
    (TAR) that uses hard temporal masks and distance-aware soft weights
    to stabilize local frame representations and reduce pseudo-label
    switching.

    \item Extensive experiments on four benchmarks demonstrate
    consistent improvements in both frame-level and segment-level
    metrics, verifying the complementary benefits of SRP and TAR.
\end{itemize}

\section{Related Work}

\subsection{Unsupervised Action Segmentation}
Action segmentation has been studied with full, weak, and no action
supervision. Supervised methods use temporal convolutions, recurrent
models, or Transformers with smoothness and boundary objectives
\cite{lea2017temporal,farha2019ms,li2020ms,ahn2021refining,
ishikawa2021alleviating,yi2021asformer, li2024object}; weakly supervised methods
exploit transcripts, timestamps, or ordered action lists
\cite{behrmann2022unified,xu2024efficient}. Unsupervised approaches
instead discover latent actions from visual and temporal structure.
CTE \cite{kukleva2019unsupervised}, UDE
\cite{swetha2021unsupervised}, and VTE \cite{vidalmata2021joint}
learn temporally structured embeddings, while recent methods couple
representation learning with iterative latent assignment.

Optimal transport provides a structured alternative to independent
frame clustering. TOT \cite{kumar2022unsupervised} performs online
frame-to-action transport, and ASOT \cite{xu2024temporally} introduces
unbalanced transport with a Gromov--Wasserstein structural objective
to handle unequal durations and missing actions. SpecT-OT adopts this
backbone rather than proposing a new solver. Its focus is the
representation space underlying transport: SRP improves
frame--prototype discrimination, while TAR internalizes local temporal
regularity into frame embeddings.

\subsection{Spectral and Temporal Representation Learning}
MLPs favor low-frequency or smoothly varying functions during
optimization \cite{xu2018understanding,rahaman2019spectral,
ronen2019convergence, li2025ustc}. Fourier features broaden the modeled frequency
range through input encoding \cite{tancik2020fourier}, whereas
reparameterization changes network weights through structured bases
\cite{zagoruyko2017diracnets, zhao2025spike, zhao2026featjnd, shi2024improved}. SRP follows the latter
route by representing projector weights with fixed Fourier bases and
learnable coefficients.

Temporal regularization suppresses fragmented predictions
\cite{farha2019ms,guan2021domain,yi2021asformer}. Without labels,
methods use temporal proximity, relative position, sequential
clustering, or assignment-level structure
\cite{kukleva2019unsupervised,vidalmata2021joint,
kumar2022unsupervised,xu2024temporally, ju2025revisiting}. TAR instead regularizes
pairwise affinities directly in the embedding space, complementing the
structural constraint in the OT backbone.
\section{Methodology}

In this section, we introduce SpecT-OT, a spectral-temporal
representation learning framework for unsupervised action
segmentation. Its design follows a central requirement of action
segmentation: the learned representation should be sensitive enough
to distinguish action changes while remaining stable across
temporally coherent frames. This requirement is particularly
important for OT-based self-training, since the transport cost and the
resulting pseudo-labels are directly determined by the learned
representation space.

\begin{figure*}[t]
    \centering
    \includegraphics[width=0.97\textwidth]{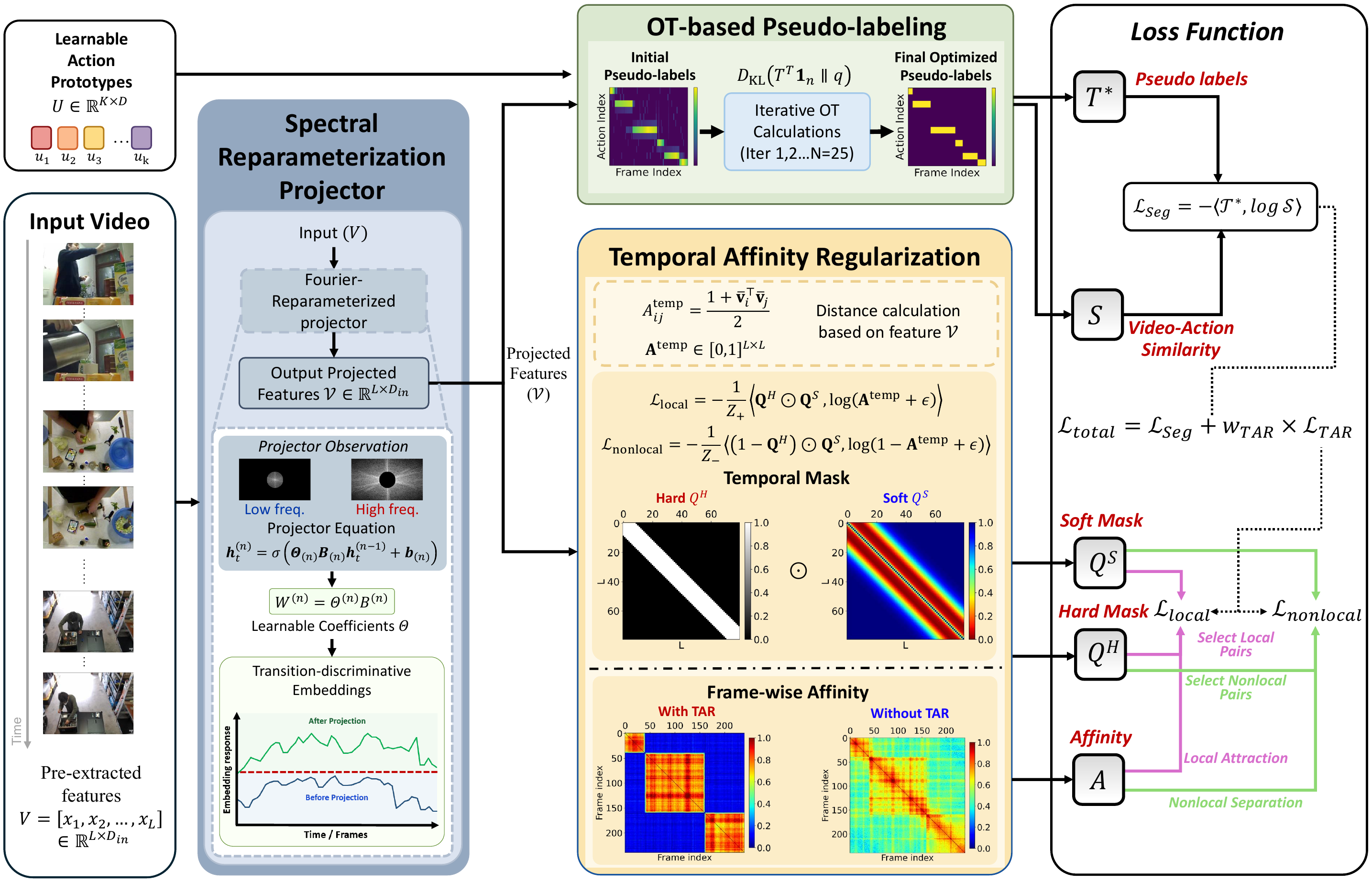}
    \vspace{-0.33cm}
    \caption{Overview of SpecT-OT. SRP represents the layer weights as 
    learnable combinations of fixed Fourier bases to discriminate feature transitions.
    Iterative unbalanced OT generates the optimized transport plan
    $\mathcal{T}^{*}$ as pseudo-labels, while TAR uses
    hard and soft temporal masks to regularize pairwise frame
    affinities. The two objectives jointly learn transition-sensitive
    and locally coherent representations.}
    \label{method structure}
\vspace{-0.5cm}
\end{figure*}

As illustrated in Fig.~\ref{method structure}, SpecT-OT builds upon
an unbalanced OT pseudo-labeling backbone and incorporates two
complementary components: the Spectral Reparameterization Projector
(SRP) and Temporal Affinity Regularization (TAR). SRP improves the
modeling of discriminative feature variations through Fourier
reparameterization, enabling the representation to better
\emph{see the change}. TAR directly regularizes pairwise frame
affinities to preserve local temporal coherence and
\emph{keep the flow}. Together, they shape a representation space
that provides more reliable costs and pseudo-labels for OT-based
learning.

\subsection{Preliminaries}

SpecT-OT aims to discover frame-wise action assignments from
pre-extracted video features without using frame-level labels,
transcripts, timestamps, or action-order annotations during training.
Following the standard protocol of unsupervised action segmentation,
the number of latent action classes $K$ is assumed to be known for
each activity.

Given a video sequence, its pre-extracted frame features are denoted as:
\begin{equation}
    V=[\mathbf{x}_{1},\ldots,\mathbf{x}_{L}]
    \in\mathbb{R}^{L\times D_{\mathrm{in}}},
\end{equation}
where $L$ is the number of frames and $D_{\mathrm{in}}$ is the input
feature dimension. As shown on the left side of
Fig.~\ref{method structure}, SRP maps these features into a latent
representation space:
\begin{equation}
    \mathcal{V}
    =
    f_{\Theta}^{\mathrm{SRP}}(V)
    =
    [\mathbf{v}_{1},\ldots,\mathbf{v}_{L}]
    \in\mathbb{R}^{L\times D},
    \label{eq:projected_features}
\end{equation}
where $\Theta$ denotes the learnable projector parameters and $D$ is
the latent feature dimension. We further maintain a learnable
action-prototype matrix:
\begin{equation}
    \mathcal{U} = [\mathbf{u}_{1},\ldots,\mathbf{u}_{K}]
    \in\mathbb{R}^{K\times D},
    \label{eq:action_prototypes}
\end{equation}
where each prototype represents a latent action category.

The projected frame embeddings and action prototypes are matched
through unbalanced OT to obtain soft frame-to-action assignments.
Importantly, OT does not operate independently of representation
learning: its frame--action cost is computed from the similarities
between $\mathcal{V}$ and $\mathcal{U}$. Ambiguous frame embeddings
therefore yield ambiguous transport costs, whereas locally unstable
embeddings may cause neighboring frames to receive inconsistent
assignments. SpecT-OT addresses these two representation-level issues
through SRP and TAR before using the resulting transport plan as
pseudo-label supervision.

\subsection{Spectral Reparameterization Projector}

To distinguish adjacent action segments, the representation space
should respond to feature variations that are informative for action
changes. However, conventional MLP projectors tend to learn smoothly
varying components more readily than rapidly varying components, a
behavior commonly referred to as spectral bias
\cite{xu2018understanding,rahaman2019spectral,
ronen2019convergence}. In action segmentation, this preference may
overemphasize persistent contextual information while weakening rapid
but discriminative variations, making frames around action
transitions difficult to separate during frame--prototype matching.

SRP addresses this limitation by reparameterizing the projector
weights over fixed Fourier bases. Rather than explicitly transforming
the video sequence into the temporal frequency domain, SRP changes the
optimization space of the projector itself. For the $n$-th projection
layer, instead of directly optimizing the weight matrix
$\mathbf{W}_{(n)}$, SRP represents it as:
\begin{equation}
\begin{aligned}
    \mathbf{W}_{(n)} &= \boldsymbol{\Theta}_{(n)}\mathbf{B}_{(n)},
    [B_{(n)}]_{(r,p),j} = \alpha\cos(\omega_r z_j+\phi_p),\\
    r &= 1,\ldots,2F,\quad p = 0,\ldots,P-1,
\end{aligned}
\label{eq:srp_weight}
\end{equation}
where $\mathbf{B}_{(n)}\in\mathbb{R}^{\gamma\times d_{n-1}}$
contains $\gamma$ fixed Fourier bases and
$\boldsymbol{\Theta}_{(n)}
\in\mathbb{R}^{d_n\times\gamma}$
contains their learnable combination coefficients. Here, $P$ denotes the number of uniformly sampled phases, and $F$
denotes the number of frequencies in each frequency bank. Specifically, the formulation is as follows:
\begin{equation}
\begin{aligned}
    \phi_p &= \frac{2\pi p}{P}, \qquad p = 0,\ldots,P-1, \\
    \omega_{\mathrm{low}} &= \left\{\frac{1}{F},\frac{2}{F},\ldots,1\right\},
    \omega_{\mathrm{high}} = \left\{1,2,\ldots,F \right\}.
\end{aligned}
\end{equation}
Combining the $P$ phases with the two frequency banks yields
$\gamma = 2F \cdot P$ fixed Fourier bases. The coordinates $z_j$ are
uniformly sampled auxiliary positions along the input-channel dimension
$d_{n-1}$.

The corresponding projection layer is formulated as follows:
\begin{equation}
    \mathbf{h}_{t}^{(n)}
    =
    \sigma\!\left(
        \boldsymbol{\Theta}_{(n)}
        \mathbf{B}_{(n)}
        \mathbf{h}_{t}^{(n-1)}
        +
        \mathbf{b}_{(n)}
    \right),
    \label{eq:srp_layer}
\end{equation}
where $\sigma(\cdot)$ denotes a nonlinear activation function. The
fixed bases provide a frequency-diverse parameterization, while the
learnable coefficients adapt their combinations to the action
segmentation objective.
The SRP architecture is controlled by the hidden feature dimension
$d_{\mathrm{hid}}$ and the number of Fourier-reparameterized hidden
layers $N_{\mathrm{FR}}$. 

SRP does not explicitly detect action boundaries or perform a
temporal Fourier transform. Instead, it improves the capacity of the
projector to model both slowly and rapidly varying feature components.
By exposing the projector to a richer spectrum of feature variations, 
SRP encourages representations that retain both discriminative 
high-frequency cues and informative low-frequency context. 
Consequently, frames exhibiting different action characteristics can
form more distinguishable similarities to the action prototypes,
producing a more informative frame--action cost matrix for OT
pseudo-labeling. In this sense, SRP enables SpecT-OT to better
\emph{see the change} in the latent representation space.

\subsection{Temporal Affinity Regularization}

Improving sensitivity to feature variations alone is insufficient for
action segmentation. Without an explicit temporal constraint, minor
appearance or motion fluctuations may be amplified and manifest as 
frequent assignment switches. Conversely, enforcing uniform
smoothness over the entire sequence may suppress genuine action
changes. A suitable temporal prior should therefore preserve local
continuity without forcing all frames to share similar
representations.

Based on this observation, TAR introduces label-free temporal
regularization directly into the embedding space. Given row-wise
normalized frame embeddings $\bar{\mathbf{v}}_i$, their pairwise
affinity $A_{ij}^{\mathrm{temp}}$ is defined as follows:
\begin{equation}
    A_{ij}^{\mathrm{temp}}
    =
    \frac{
        1+\bar{\mathbf{v}}_{i}^{\top}\bar{\mathbf{v}}_{j}
    }{2},
    \qquad
    \mathbf{A}^{\mathrm{temp}}\in[0,1]^{L\times L}.
    \label{eq:frame_affinity}
\end{equation}

TAR relies on a weak but generally valid temporal prior: nearby frames
are more likely to belong to the same action progression than distant
frames. As shown in the TAR block of Fig.~\ref{method structure}, a
hard temporal mask $Q^{H}$ first separates local and nonlocal frame pairs:
\begin{equation}
    Q^{H}_{ij}
    =
    \begin{cases}
        1, & |i-j|\leq\delta,\\
        0, & |i-j|>\delta,
    \end{cases}
    \label{eq:hard_mask}
\end{equation}
where $\delta$ controls the temporal-window size. Since frame pairs within
the same group should not contribute equally, a soft mask $Q^{S}$ further
weights them according to temporal distance:
\begin{equation}
    Q^{S}_{ij}
    =
    \begin{cases}
        \dfrac{1}{(i-j)^2+1},
        & |i-j|\leq\delta,\\[5pt]
        1-\exp\!\left(
            -\dfrac{(|i-j|-\delta)^2}{\varsigma}
        \right),
        & |i-j|>\delta,
    \end{cases}
    \label{eq:soft_mask}
\end{equation}
where $\varsigma$ controls the nonlocal weighting scale.

The two masks play complementary roles. The hard mask prevents
temporal regularization from indiscriminately smoothing the entire
sequence, while the soft mask modulates the constraint strength:
closer frames receive stronger coherence constraints, whereas
increasingly distant frames receive stronger separation constraints.
The local and nonlocal affinity objectives are formulated as follows:
\begin{equation}
\begin{aligned}
    \mathcal{L}_{\mathrm{local}}
    &=
    -\frac{1}{Z_{+}}
    \left\langle
        \mathbf{Q}^{H}\odot\mathbf{Q}^{S},
        \log(\mathbf{A}^{\mathrm{temp}}+\epsilon)
    \right\rangle,\\
    \mathcal{L}_{\mathrm{nonlocal}}
    &=
    -\frac{1}{Z_{-}}
    \left\langle
        (1-\mathbf{Q}^{H})\odot\mathbf{Q}^{S},
        \log(1-\mathbf{A}^{\mathrm{temp}}+\epsilon)
    \right\rangle,
\end{aligned}
\label{eq:tar_components}
\end{equation}
where $Z_{+}$ and $Z_{-}$ normalize the corresponding mask weights,
$\odot$ denotes element-wise multiplication, and $\epsilon$ ensures
numerical stability. The complete TAR objective is
\begin{equation}
    \mathcal{L}_{\mathrm{TAR}}
    =
    \mathcal{L}_{\mathrm{local}}
    +
    \lambda_{\mathrm{neg}}
    \mathcal{L}_{\mathrm{nonlocal}},
    \label{eq:tar_loss}
\end{equation}
where $\lambda_{\mathrm{neg}}$ balances local attraction and nonlocal
separation.

The local objective encourages neighboring frames to form stable
affinity structures, while the nonlocal objective prevents all
embeddings from collapsing into a single representation. 
TAR does not merely smooth the final predictions; it
internalizes temporal regularity into the representation space from
which the OT costs are constructed. As visualized in
Fig.~\ref{method structure}, TAR produces clearer locally coherent
affinity structures and reduces fragmented assignments. In this way,
TAR allows SpecT-OT to \emph{keep the flow} without erasing meaningful
action changes.

\subsection{OT-Based Pseudo-Label Learning}

The unbalanced OT backbone addresses the assignment-side mismatch
caused by unequal action durations and missing actions. However, its
solution still depends on the geometry supplied by the projected frame
embeddings and action prototypes. SpecT-OT therefore does not
introduce a new OT solver; instead, SRP and TAR improve the
representation space from which the transport problem is constructed.

The projected frame embeddings and action prototypes are first
$\ell_2$-normalized. Their frame--action transport cost is defined as:
\begin{equation}
    C_{lk} = 1- \bar{\mathbf{v}}_{l}^{\top}\bar{\mathbf{u}}_{k},
    \qquad
    \mathbf{C}\in\mathbb{R}^{L\times K}.
    \label{eq:transport_cost}
\end{equation}
Following the unbalanced OT formulation
\cite{xu2024temporally}, the frame-side marginal $\mathbf{p}$ is
preserved, whereas the action-side marginal $\mathbf{q}$ is softly
constrained:
\begin{equation}
\begin{aligned}
    \mathcal{T}^{*}
    =
    \arg\min_{
        \mathcal{T}\geq0,\,
        \mathcal{T}\mathbf{1}_{K}=\mathbf{p}}
    \;&
    \langle\mathbf{C},\mathcal{T}\rangle
    +
    \alpha\,
    \Omega_{\mathrm{str}}
    (\mathcal{T};\mathbf{C}^{p},\mathbf{C}^{q})
    \\
    &+
    \lambda\,
    D_{\mathrm{KL}}\!\left(
        \mathcal{T}^{\top}\mathbf{1}_{L}
        \,\|\,\mathbf{q}
    \right),
\end{aligned}
\label{eq:uot}
\end{equation}
where $\mathbf{C}^{p}$ and $\mathbf{C}^{q}$ denote the intra-set
costs illustrated in Fig.~\ref{method structure}, and
$\Omega_{\mathrm{str}}$ is the structural temporal objective of the
OT backbone. Relaxing the action-side marginal allows different
action prototypes to receive unequal transport mass and accommodates
actions that are absent from individual videos. The optimized
transport plan
$\mathcal{T}^{*}\in\mathbb{R}_{+}^{L\times K}$ is detached from
gradient computation and used as soft pseudo-label supervision.
The frame--prototype similarity is computed between normalized features:
$s_{lk} = softmax (\bar{\mathbf{v}}_{l}^{\top}\bar{\mathbf{u}}_{k}/\tau)$,
where $\tau$ is a temperature parameter. The pseudo-label
segmentation loss is formulated as follows:
\begin{equation}
    \mathcal{L}_{\mathrm{Seg}}
    =
    -\frac{1}{BL}
    \sum_{b=1}^{B}
    \sum_{l=1}^{L}
    \sum_{k=1}^{K}
    \operatorname{sg}(t_{lk}^{*b})
    \log s_{lk}^{b},
    \label{eq:seg_loss}
\end{equation}
where $B$ denotes the batch size and
$\operatorname{sg}(\cdot)$ denotes the stop-gradient operation. Finally, the overall
training objective is formulated as follows:
\begin{equation}
    \mathcal{L}_{\mathrm{total}}
    =
    \mathcal{L}_{\mathrm{Seg}}
    +
    w_{\mathrm{TAR}}
    \mathcal{L}_{\mathrm{TAR}},
    \label{final_loss}
\end{equation}
where $w_{\mathrm{TAR}}$ controls the strength of temporal affinity
regularization.

The three components form a coupled self-training process. SRP
improves the discrimination of frame--prototype costs, TAR stabilizes
their local temporal geometry, and unbalanced OT converts the
resulting representation structure into soft pseudo-labels. These
pseudo-labels in turn supervise the projector and action prototypes.
Thus, SRP and TAR jointly address
the two complementary requirements of action segmentation: being
sensitive to action changes and coherent along local temporal
progressions.

\section{Experiments}
\label{experimental}

\subsection{Experiment Settings}
\label{setup}
\noindent\textbf{Datasets.}
We evaluate SpecT-OT on Breakfast (BF) \cite{kuehne2014language},
50 Salads (FS) \cite{stein2013combining}, YouTube Instructions (YTI)
\cite{alayrac2016unsupervised}, and Desktop Assembly (DA)
\cite{kumar2022unsupervised}. BF contains about 1,700 videos, 10
activities, and 48 actions; FS contains 50 videos and is evaluated under two different annotation granularities: Mid (19 classes) and Eval (12 classes) ; YTI contains 150
videos from five activities with substantial background; and DA
contains 76 videos with 22 fine-grained actions. Following prior work
\cite{kukleva2019unsupervised,kumar2022unsupervised,li2021action},
we use publicly released pre-extracted features.

\noindent\textbf{Implementation.}
Action prototypes are initialized by K-means clustering \cite{caron2018deep}.
We use Adam with learning rate $10^{-3}$, weight decay $10^{-4}$,
25 OT iterations, and temporal window $\delta=3$. SRP uses $P=16$ phase samples and $F=64$ frequencies
in each of the low- and high-frequency banks, resulting in
$\gamma = 2F \cdot P = 2048$ Fourier bases. The tuples
$(d_{\mathrm{hid}},N_{\mathrm{FR}},w_{\mathrm{TAR}})$ are
$(256,2,10^{-4})$, $(256,8,10^{-3})$,
$(128,2,10^{-3})$, $(16,2,10^{-3})$, and
$(1024,1,10^{-4})$ for FS-Mid, FS-Eval, BF, YTI, and DA, respectively.
All experiments run on a single RTX 4090 GPU.

\noindent\textbf{Evaluation.}
Hungarian matching aligns discovered clusters with ground-truth
action labels. We report \emph{full} matching, which shares one mapping across
videos of the same activity, and \emph{per-video} matching, which
matches each video independently \cite{xu2024temporally}. Evaluation metrics
include MoF, segmental F1 at $50\%$ overlap, and mIoU. SpecT-OT uses
neither action-order supervision nor HMM post-processing.

\begin{table*}[t]
\centering
\caption{Comparison under full and per-video Hungarian matching.
Best and second-best results within each matching protocol are shown
in bold and underlined, respectively; ``--'' denotes unavailable
results. Methods include a publication-year suffix. 
The light-blue highlight our method, SpecT-OT.
Notation $^{\dagger}$ denotes reproduced results.
Notation $^{\ddagger}$ denotes an adapted reproduction of MulSclTE using the
same pre-extracted input features as ASOT and SpecT-OT.}
\vspace{-0.25cm}
\label{tab:main_results}

\fontsize{8}{8}\selectfont
\setlength{\tabcolsep}{0.4pt}
\renewcommand{\arraystretch}{1}
\def\rot#1{\rotatebox[origin=c]{75}{#1}}

\begin{tabular*}{\textwidth}{
@{\extracolsep{\fill}}
ll
*{13}{c}
>{\columncolor{ourcfg}}c|
*{5}{c}
>{\columncolor{ourcfg}}c
@{}
}
\toprule
\multirow{2}{*}{Data}
& \multirow{2}{*}{Metric}
& \multicolumn{14}{c|}{\emph{Full} Matching}
& \multicolumn{6}{c}{\emph{Per-video} Matching} \\
\cmidrule(lr){3-16}
\cmidrule(lr){17-22}

&
& \rot{CTE (2019)}
& \rot{VTE (2021)}
& \rot{UDE (2021)}
& \rot{ASAL (2021)}
& \rot{TOT (2022)}
& \rot{TOT+TCL (2022)}
& \rot{UFSA (2024)}
& \rot{ASOT$^{\dagger}$ (2024)}
& \rot{VASOT$^{\dagger}$ (2025)}
& \rot{HVQ$^{\dagger}$ (2025)}
& \rot{CLOT$^{\dagger}$ (2025)}
& \rot{MulSclTE$^{\ddagger}$ (2025)}
& \rot{PEOT$^{\dagger}$ (2026)}
& \rot{\textbf{SpecT-OT}}
& \rot{TWF (2021)}
& \rot{ABD (2022)}
& \rot{ASOT$^{\dagger}$ (2024)}
& \rot{VASOT$^{\dagger}$ (2025)}
& \rot{CLOT$^{\dagger}$ (2025)}
& \rot{\textbf{SpecT-OT}} \\
\midrule

\multirow{3}{*}{\shortstack{FS \\(Mid)}}
& MoF
& 30.0 & 24.2 & 31.3 & 34.4 & 31.8 & 34.3 & 36.7 & 46.6
& \underline{48.1} & -- & 31.5 & 14.4 & -- & \textbf{48.8}
& 66.8 & \textbf{71.8} & 65.6 & \underline{67.4} & 43.5 & 67.3 \\

& F1
& 25.8 & -- & 24.1 & -- & -- & -- & 30.4 & 38.2
& \underline{41.8} & -- & 21.6 & 10.4 & -- & \textbf{41.9}
& \underline{56.4} & -- & 51.6 & \textbf{61.1} & 27.8
& \textbf{61.1} \\

& mIoU
& 18.1 & -- & -- & -- & -- & -- & -- & 25.2
& \textbf{28.4} & -- & 12.8 & 3.6 & -- & \underline{27.6}
& \textbf{48.7} & -- & 33.7 & \underline{43.6} & 17.3 & 41.5 \\
\midrule

\multirow{3}{*}{\shortstack{FS \\(Eval)}}
& MoF
& 35.3 & 30.6 & 42.1 & 39.2 & 47.4 & 44.5
& \underline{55.8} & 54.4 & 44.9 & -- & 41.6 & 14.4 & 49.5
& \textbf{57.9}
& \textbf{71.7} & \underline{71.2} & 59.8 & 56.5 & 53.1 & 62.9 \\

& F1
& 35.5 & -- & 34.6 & -- & 42.8 & 48.2 & 50.3
& \underline{52.4} & 47.7 & -- & 42.4 & 10.4 & 41.9
& \textbf{54.1}
& -- & -- & \underline{56.4} & 56.0 & 52.9 & \textbf{58.7} \\

& mIoU
& 21.4 & -- & -- & -- & -- & -- & --
& \underline{29.0} & 24.9 & -- & 26.4 & 3.6 & 12.2
& \textbf{29.9}
& -- & -- & 31.6 & \underline{32.8} & 31.0 & \textbf{33.6} \\
\midrule

\multirow{3}{*}{BF}
& MoF
& 47.1 & 48.1 & 47.4 & 52.5 & 47.5 & 39.0 & 52.1 & 52.9
& 55.4 & 44.1 & 51.8 & 11.8 & \underline{56.3} & \textbf{57.0}
& 62.7 & 64.0 & 61.8 & \underline{64.5} & 60.6 & \textbf{65.4} \\

& F1
& 27.2 & -- & 32.0 & 37.9 & 31.0 & 30.3
& \underline{38.0} & 36.9 & 37.3 & 32.6 & 37.5 & 1.6
& \textbf{38.5} & 37.8
& 49.8 & 52.3 & 52.6 & \underline{54.3} & \textbf{56.6} & 53.5 \\

& mIoU
& 15.0 & -- & -- & -- & -- & -- & -- & 17.4
& 17.7 & -- & \textbf{19.0} & 1.0 & 18.6 & \underline{18.9}
& \textbf{42.3} & -- & 34.1 & 34.6 & \underline{37.3} & 36.1 \\
\midrule

\multirow{3}{*}{YTI}
& MoF
& 35.5 & -- & 43.8 & 44.9 & 40.6 & 45.3 & 49.6
& \underline{50.2} & 46.6 & 23.8 & 47.2 & 9.4 & 48.3
& \textbf{53.1}
& 56.7 & 67.2 & \textbf{70.1} & \underline{69.7} & 67.3 & 68.7 \\

& F1
& 28.5 & 29.9 & 29.6 & 32.1 & 30.0 & 32.9 & 32.4
& 34.3 & 34.2 & 15.8 & 33.6 & 0.8 & \underline{35.0}
& \textbf{36.6}
& 48.2 & 49.2 & \underline{62.9} & \textbf{63.6} & 58.6 & 59.7 \\

& mIoU
& 9.9 & -- & -- & -- & -- & -- & --
& \textbf{24.2} & 22.8 & -- & 19.7 & 1.1 & \underline{23.5}
& \underline{23.5}
& -- & -- & \underline{47.5} & \textbf{48.3} & 42.2 & 43.2 \\
\midrule

\multirow{3}{*}{\shortstack{DA}}
& MoF
& 48.1 & -- & 49.1 & -- & 56.3 & 58.1
& \underline{65.4} & 62.5 & 63.1 & -- & 41.2 & -- & 63.5
& \textbf{65.5}
& \textbf{73.3} & -- & 69.5 & 70.0 & 53.2 & \underline{71.6} \\

& F1
& 45.0 & -- & 48.3 & -- & 51.7 & 53.4 & 63.0 & 61.4
& \underline{64.4} & -- & 35.1 & -- & 61.9 & \textbf{68.8}
& \underline{67.7} & -- & 64.3 & \textbf{73.8} & 47.2
& \textbf{73.8} \\

& mIoU
& 19.4 & -- & -- & -- & -- & -- & -- & 40.2
& \underline{45.0} & -- & 22.5 & -- & 40.1 & \textbf{45.3}
& \textbf{57.7} & -- & 43.8 & \underline{51.4} & 31.5 & 50.8 \\
\bottomrule
\end{tabular*}
\vspace{-0.45cm}
\end{table*}

\begin{table*}[t]
\centering

\begin{tabular}[t]{@{}
    p{0.55\textwidth}
    p{0.45\textwidth}
    @{}}

\vspace{0pt}
\centering

\captionof{table}{Ablation of SRP and TAR under full Hungarian matching.}
\label{module_ablation}
\vspace{-0.25cm}

\fontsize{8}{8.5}\selectfont
\setlength{\tabcolsep}{0.75pt}
\renewcommand{\arraystretch}{1.15}

\begin{tabular}{@{}l*{15}{c}@{}}
\toprule
& \multicolumn{3}{c}{\shortstack{50 Salads\\(Mid)}}
& \multicolumn{3}{c}{\shortstack{50 Salads\\(Eval)}}
& \multicolumn{3}{c}{Breakfast}
& \multicolumn{3}{c}{YouTube}
& \multicolumn{3}{c}{\shortstack{Desktop\\Assembly}} \\
\cmidrule(lr){2-4}
\cmidrule(lr){5-7}
\cmidrule(lr){8-10}
\cmidrule(lr){11-13}
\cmidrule(lr){14-16}

Setting
& MoF & F1 & mIoU
& MoF & F1 & mIoU
& MoF & F1 & mIoU
& MoF & F1 & mIoU
& MoF & F1 & mIoU \\
\midrule

\makecell[l]{w/o SRP\\and TAR}
& 45.2 & 36.9 & 24.6
& 52.0 & 49.8 & 28.0
& 50.8 & 35.0 & 16.9
& 49.2 & 34.3 & 23.0
& 57.5 & 56.4 & 35.2 \\
\specialrule{0.3pt}{0pt}{0pt}

SRP only
& 47.4 & 41.1 & 27.0
& 52.7 & 53.0 & 28.1
& 56.4 & 35.9 & 18.1
& 52.1 & 36.0 & 22.3
& 60.6 & 64.3 & 41.8 \\
\specialrule{0.3pt}{0pt}{0pt}

TAR only
& 47.2 & 38.6 & 26.5
& 57.1 & 53.2 & 30.0
& 53.5 & 36.1 & 17.7
& 51.2 & 35.5 & 22.7
& 63.1 & 66.7 & 44.5 \\
\specialrule{0.3pt}{0pt}{0pt}

\rowcolor{ourcfg}
\multicolumn{1}{
  @{}
  >{\columncolor{ourcfg}[0.5pt][7pt]}
  c
}{
  \makecell[c]{SpecT-\\OT}
  }
& 48.8 & 41.9 & 27.6
& 57.9 & 54.1 & 29.9
& 57.0 & 37.8 & 18.9
& 53.1 & 36.6 & 23.5
& 65.5 & 68.8 & 45.3 \\
\bottomrule
\end{tabular}

&

\vspace{0pt}
\centering

\captionof{table}{Runtime and computational cost.}
\label{time_and_memory}
\vspace{-0.25cm}

\fontsize{8}{8.5}\selectfont
\setlength{\tabcolsep}{1.15pt}
\renewcommand{\arraystretch}{0.95}

\begin{tabular}{@{}llccccc@{}}
\toprule
Metric
& Model
& \shortstack{FS (Mid)}
& \shortstack{FS (Eval)}
& BF
& YTI
& \shortstack{DA} \\
\midrule

\multirow{2}{*}{GFLOPs}
& ASOT
& 0.31 & 0.26 & 0.88 & 0.80 & 1.58 \\
& \cellcolor{ourcfg}SpecT-OT
& \cellcolor{ourcfg}24.75
& \cellcolor{ourcfg}107.90
& \cellcolor{ourcfg}92.36
& \cellcolor{ourcfg}1.86
& \cellcolor{ourcfg}329.57 \\
\midrule

\multirow{2}{*}{\shortstack[l]{Memory \\ (GB)}}
& ASOT
& 6.71 & 6.12 & 5.47 & 5.12 & 5.59 \\
& \cellcolor{ourcfg}SpecT-OT
& \cellcolor{ourcfg}6.24
& \cellcolor{ourcfg}7.16
& \cellcolor{ourcfg}5.89
& \cellcolor{ourcfg}5.47
& \cellcolor{ourcfg}7.76 \\
\midrule

\multirow{2}{*}{\shortstack[l]{Time/\\epoch (s)}}
& ASOT
& 13.51 & 13.32 & 21.11 & 9.07 & 4.98 \\
& \cellcolor{ourcfg}SpecT-OT
& \cellcolor{ourcfg}14.10
& \cellcolor{ourcfg}13.74
& \cellcolor{ourcfg}16.49
& \cellcolor{ourcfg}8.82
& \cellcolor{ourcfg}5.53 \\
\midrule

\multirow{2}{*}{Epochs}
& ASOT
& 30 & 30 & 15 & 10 & 30 \\
& \cellcolor{ourcfg}SpecT-OT
& \cellcolor{ourcfg}140
& \cellcolor{ourcfg}80
& \cellcolor{ourcfg}25
& \cellcolor{ourcfg}10
& \cellcolor{ourcfg}100 \\
\bottomrule
\end{tabular}

\end{tabular}
\vspace{-0.6cm}
\end{table*}

\subsection{Comparison with State of the Art}
Table~\ref{tab:main_results} compares SpecT-OT with representative
unsupervised action segmentation methods published from year 2019 to 2026.
We used open source results for earlier baselines and reproduce five
recent methods, namely ASOT \cite{xu2024temporally}, VASOT \cite{ali2025joint}, HVQ \cite{hvq2025spurio}, CLOT \cite{bueno2025clot}, and PEOT \cite{li2026learning}. 
Their reproduced results are marked with $^{\dagger}$. 
For MulSclTE \cite{song2025unsupervised}, whose official implementation uses 2048-dimensional
frame-wise I3D features, we replace the original input with the same
pre-extracted features used by ASOT and SpecT-OT, adjusting only the
input layer to accommodate the feature dimension while retaining all
subsequent network components. Therefore, its results are marked with
$^{\ddagger}$ to denote an adapted reproduction under a common
input-feature setting.

Under full matching, SpecT-OT ranks first on 11 of 15 metrics.
Relative to ASOT, it improves Breakfast MoF by 4.1 points and
Desktop Assembly F1/mIoU by 7.4/5.1 points; only Breakfast F1 and
YTI mIoU remain slightly below the best results. 
These results indicate that improving the representation space
benefits both frame-level recognition and segment-level consistency
under the more restrictive dataset-level matching protocol.

Under per-video matching, SpecT-OT obtains the best or tied-best
result on five metrics and the second-best result on one metric. It
achieves particularly strong segmental F1 on FS-Mid, FS-Eval, and
Desktop Assembly, as well as the best MoF on Breakfast. Performance
on YTI is less competitive under per-video matching, indicating that
the improvements observed under dataset-level matching do not
uniformly transfer to video-specific cluster alignment.

\subsection{Qualitative Analysis}

\begin{figure*}[t]
\centering
\captionsetup[subfloat]{labelfont=footnotesize,textfont=footnotesize}

\subfloat[50 Salads]{%
    \includegraphics[width=0.31\textwidth]{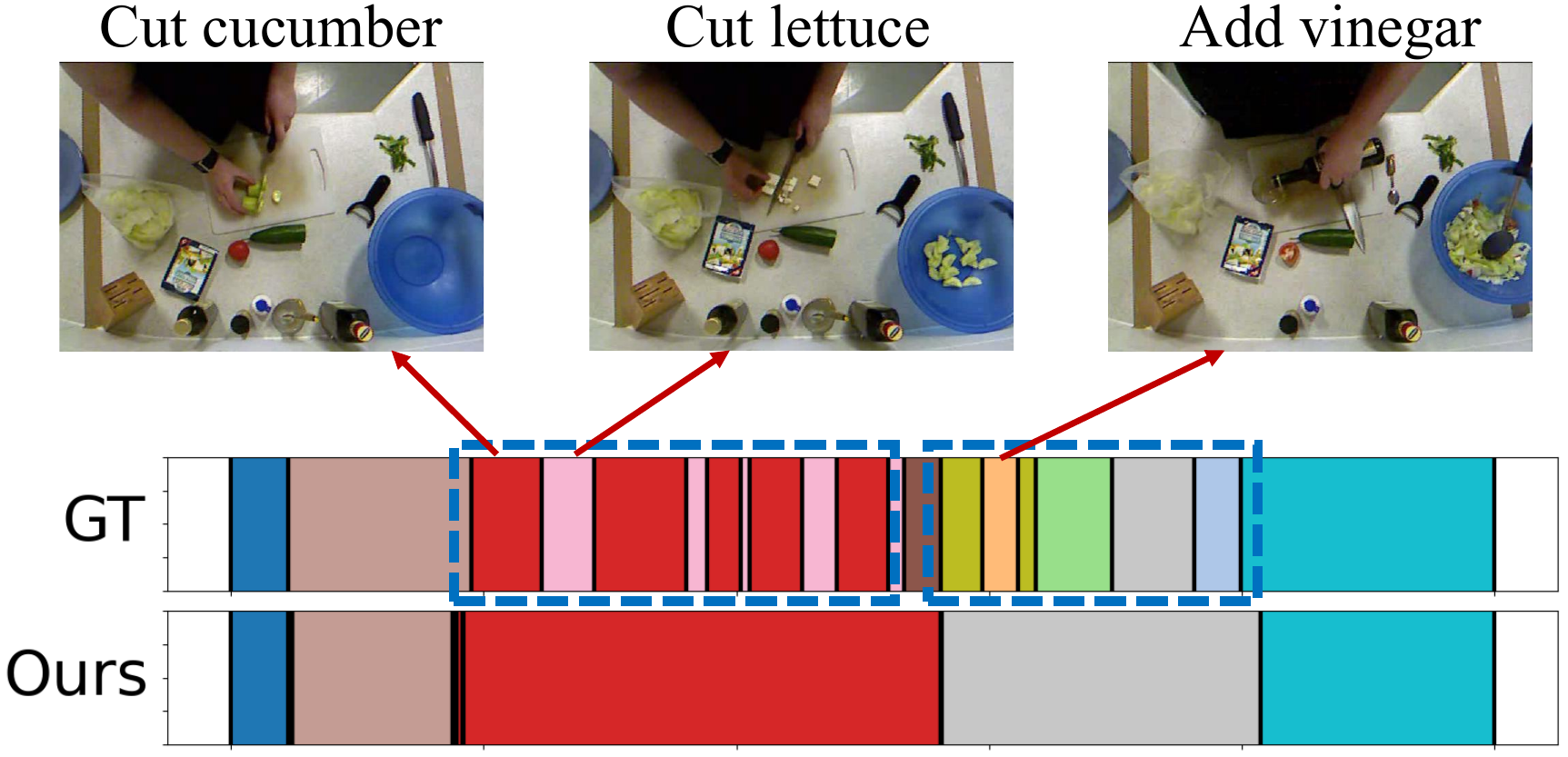}%
    \label{50_salads_25-1}}
\hfill
\subfloat[Breakfast]{%
    \includegraphics[width=0.31\textwidth]{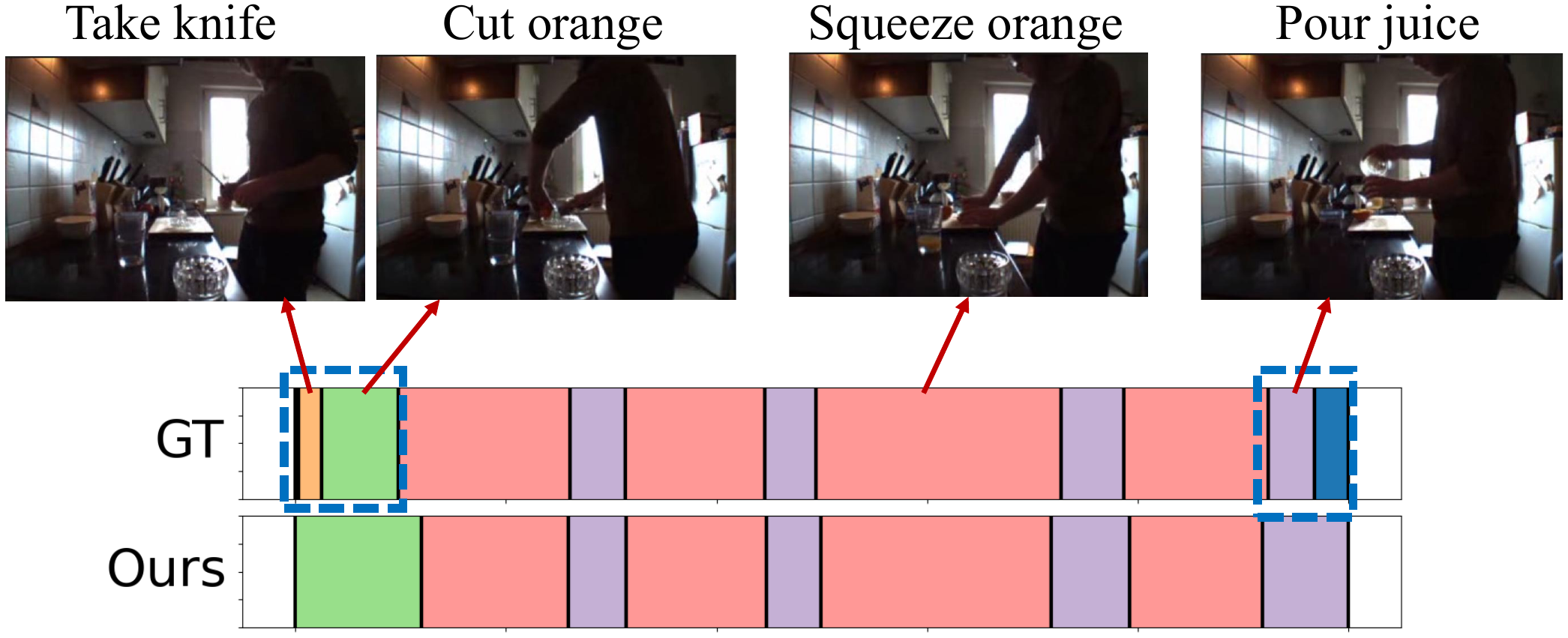}%
    \label{visual_Breakfast}}
\hfill
\subfloat[Desktop Assembly]{%
    \includegraphics[width=0.31\textwidth]{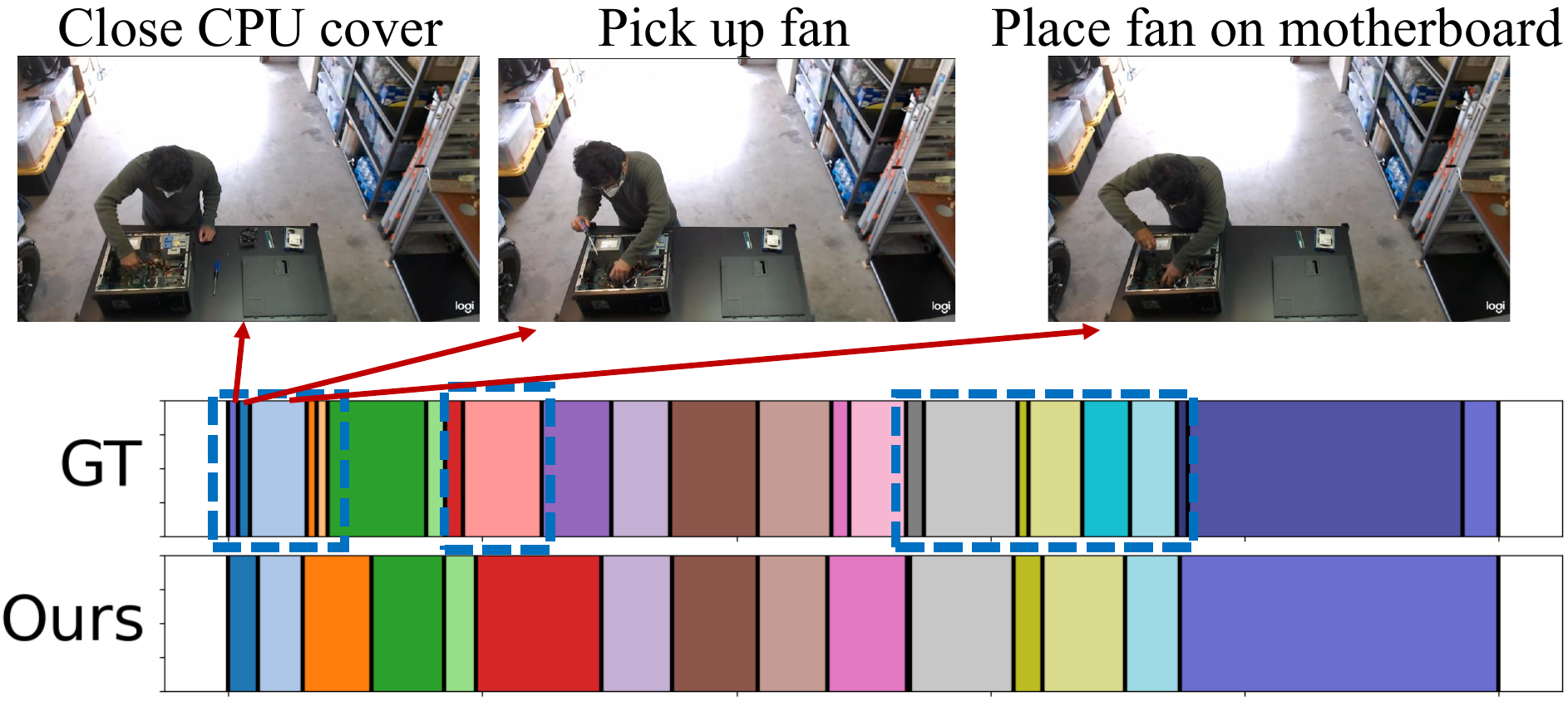}%
    \label{visual_Desktop}}
\vspace{-0.2cm}
\caption{Qualitative segmentation results. Colored regions denote
predicted action segments, while blue boxes highlight rapid action
changes or representative errors.}
\label{visual_qualitative}
\vspace{-0.3cm}
\end{figure*}

Fig.~\ref{visual_qualitative} compares the predicted segmentations
with the ground truth on three representative datasets. Overall,
SpecT-OT produces temporally coherent action regions and avoids
frequent label switching within long action segments. At the same
time, most major action transitions remain distinguishable, showing
that improving local coherence does not simply reduce the predictions
to uniformly smoothed segments.

The examples also reveal the remaining challenges. On 50 Salads,
SpecT-OT correctly preserves the long action regions surrounding the
highlighted interval, but some short object-specific cutting and
dressing actions within the densely changing region are merged. On
Breakfast, the major action intervals are well aligned, whereas
confusion remains around the transition from \emph{take knife} to
\emph{cut orange}, whose neighboring frames contain highly similar
visual content. On Desktop Assembly, the model generates stable
predictions for long assembly stages but may shift or omit very short
operations involving the CPU cover and fan.

\begin{figure*}[t]
\centering

\captionsetup[subfloat]{
    labelfont=footnotesize,
    textfont=footnotesize
}

\subfloat[SRP hidden dimension $d_{hid}$]{%
    \includegraphics[width=0.32\textwidth]
    {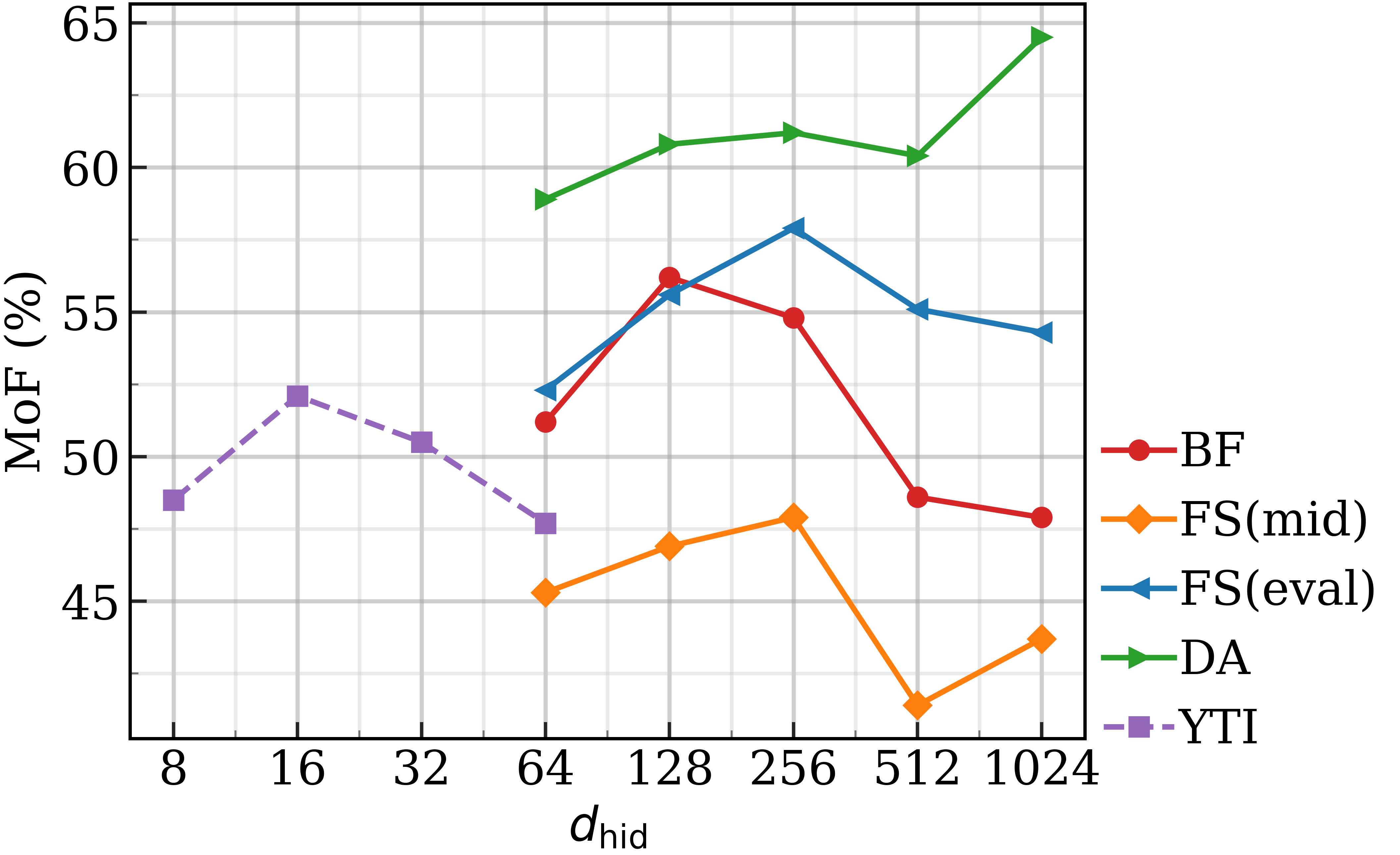}%
    \label{Ablation_features}%
}
\hfill
\subfloat[SRP hidden layers $N_{FR}$]{%
    \includegraphics[width=0.32\textwidth]
    {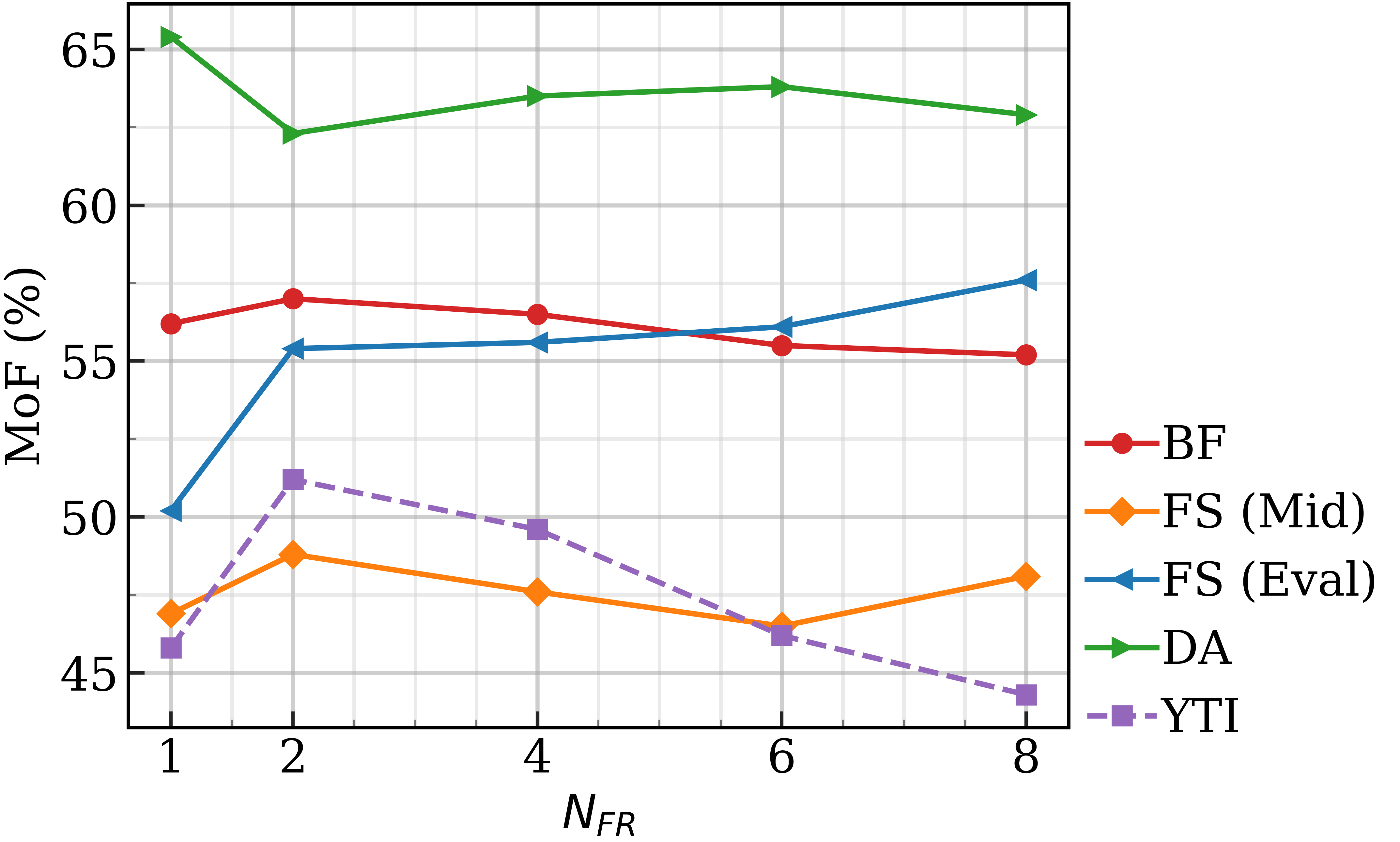}%
    \label{Ablation_layers}%
}
\hfill
\subfloat[TAR weight $w_{\mathrm{TAR}}$]{%
    \includegraphics[width=0.32\textwidth]
    {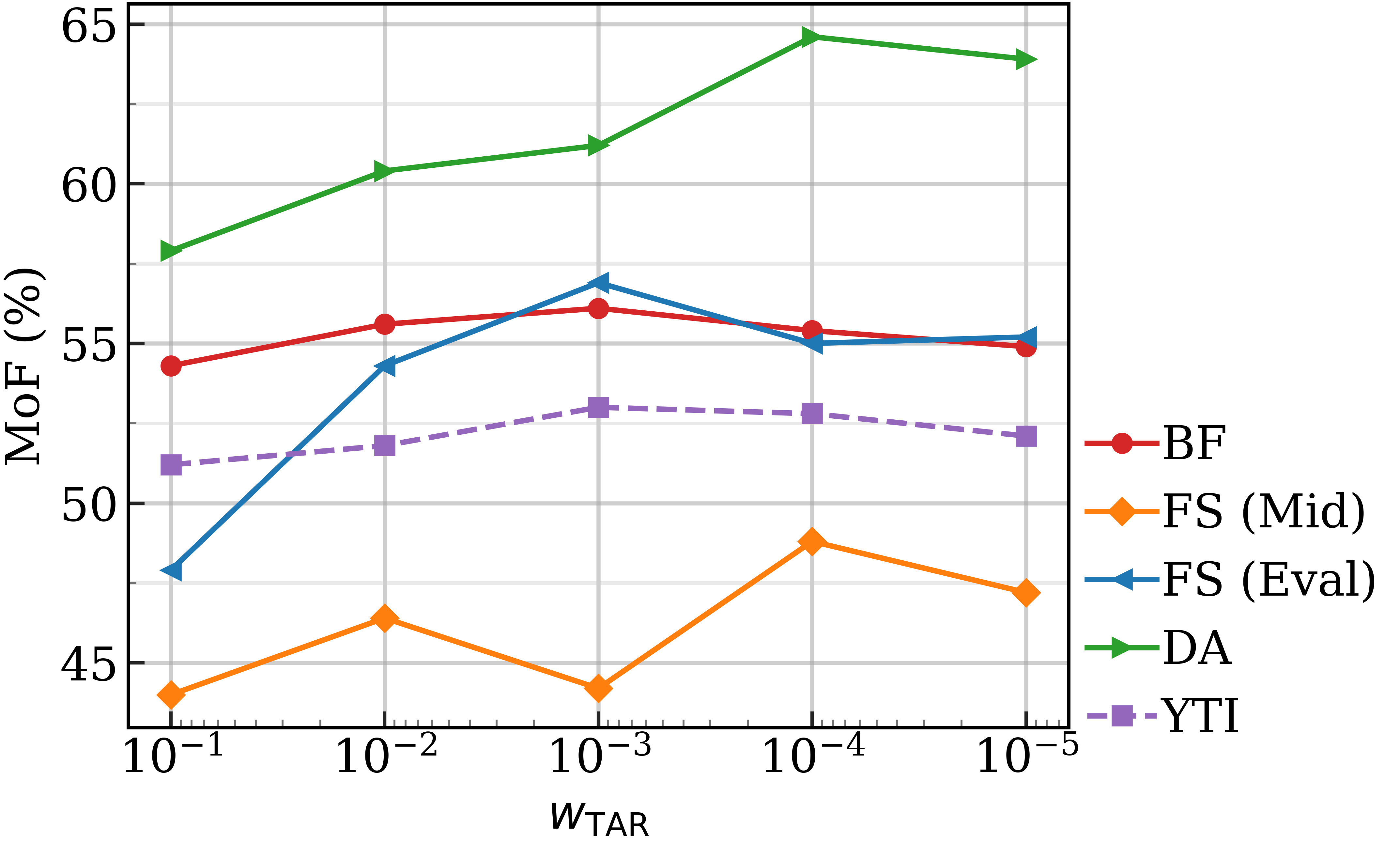}%
    \label{Ablation_weight_tc}%
}
\vspace{-0.2cm}
\caption{Sensitivity to the SRP hidden dimension
$d_{\mathrm{hid}}$, the Fourier-reparameterized hidden
layers $N_{\mathrm{FR}}$, and the TAR loss weight
$w_{\mathrm{TAR}}$. All results are reported using MoF.}
\label{hyperparameter_analysis}
\vspace{-0.5cm}
\end{figure*}

\begin{figure}[t]
    \centering
    \includegraphics[width=0.35\textwidth]{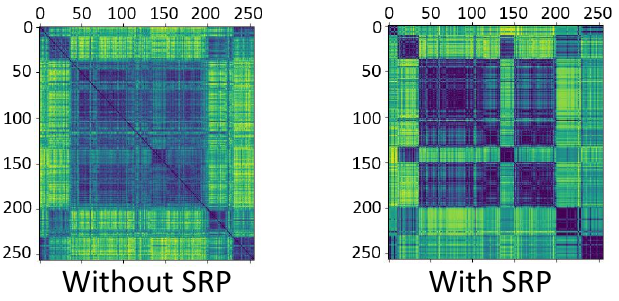}
    \vspace{-0.25cm}
    \caption{Frame-wise cosine-distance matrices with and
    without SRP. Darker color indicate greater embedding similarity.}
\label{Ablation_INR}
\vspace{-0.5cm}
\end{figure}

\subsection{Ablation Studies}
\label{ablation}

Table~\ref{module_ablation} verifies the complementary roles of SRP
and TAR. SRP improves 14 of 15 entries over the vanilla
MLP projector, with an average gain of 2.8 points, while TAR also
improves 14 entries with an average gain of 3.3 points. Combining the
two modules improves all 15 entries over the baseline by an average of
4.8 points and outperforms the stronger single-module variant on
14 entries. The only exception is FS-Eval mIoU, where TAR alone is
0.1 point higher. These results support the complementary roles of
transition-discriminative projection and local affinity regularization.


Fig.~\ref{hyperparameter_analysis} analyzes the SRP architecture and
TAR weight. The optimal projector width is dataset dependent, while
one or two Fourier-reparameterized layers are sufficient for most
datasets. Increasing the width or depth does not consistently improve
performance. For TAR, the most reliable results are obtained with
$w_{\mathrm{TAR}}$ between $10^{-4}$ and $10^{-3}$. A large weight such
as $10^{-1}$ overemphasizes the temporal prior and can suppress useful
discriminative variations, whereas an excessively small weight weakens
its regularization effect. Fig.~\ref{Ablation_INR} shows a visualization of frame-wise cosine-distance. 
The clearer block-diagonal affinity with SRP indicates improved separation while preserving within-segment similarity.

\subsection{Efficiency Analysis}

Table~\ref{time_and_memory} compares SpecT-OT with ASOT on a single
RTX 4090 GPU. SpecT-OT uses $5.47$--$7.76$\,GB of memory, remaining
within a practical single-GPU budget. Although its nominal GFLOPs are
higher, the measured per-epoch runtime changes only slightly and is
even faster on Breakfast and YTI, indicating that the parallel
matrix operations introduced by SRP do not translate proportionally
into wall-clock overhead.
The main additional cost comes from the larger number of training
epochs, particularly on FS-Mid and Desktop Assembly. Nevertheless,
SpecT-OT maintains moderate per-epoch runtime and a memory footprint
below $8$\,GB, making the proposed representation learning framework
computationally feasible on a single GPU.

\section{Conclusion}

In this paper, we propose SpecT-OT, a spectral-temporal
representation learning framework for unsupervised action
segmentation. SpecT-OT formulates the representation space
as a transition-discriminative generation through
two complementary components. The SRP represents projector weights
using Fourier bases and learnable coefficients, thus
improving the modeling of discriminative feature variations
associated with action changes. TAR further imposes label-free, distance-aware
constraints on pairwise frame affinities to preserve local
temporal coherence and reduce unstable pseudo-label
switching. Extensive experiments on four public benchmarks
demonstrate the effectiveness and complementary of the
two components. In particular, SpecT-OT achieves the best
results on 11 of 15 metrics across the datasets.
Together, they enable the model to \emph{see the change} and \emph{keep the flow}, yielding consistent improvements under both full and per-video matching.

\bibliographystyle{IEEEtran}
\bibliography{refs}

@inproceedings{xu2024temporally,
  title={Temporally Consistent Unbalanced Optimal Transport for Unsupervised Action Segmentation},
  author={Xu, Ming and Gould, Stephen},
  booktitle={Proceedings of the IEEE Conference on Computer Vision and Pattern Recognition},
  pages={14618--14627},
  year={2024}
}

@article{shi2024improved,
  title={Improved implicit neural representation with fourier reparameterized training},
  author={Shi, Kexuan and Zhou, Xingyu and Gu, Shuhang},
  journal={Proceedings of the IEEE Conference on Computer Vision and Pattern Recognition},
  pages={25985--25994},
  year={2024}
}

@inproceedings{kumar2022unsupervised,
  title={Unsupervised action segmentation by joint representation learning and online clustering},
  author={Kumar, Sateesh and Haresh, Sanjay and Ahmed, Awais and Konin, Andrey and Zia, M Zeeshan and Tran, Quoc-Huy},
  booktitle={Proceedings of the IEEE Conference on Computer Vision and Pattern Recognition},
  pages={20174--20185},
  year={2022}
}

@inproceedings{kuehne2014language,
  title={The language of actions: Recovering the syntax and semantics of goal-directed human activities},
  author={Kuehne, Hilde and Arslan, Ali and Serre, Thomas},
  booktitle={Proceedings of the IEEE Conference on Computer Vision and Pattern Recognition},
  pages={780--787},
  year={2014}
}

@inproceedings{stein2013combining,
  title={Combining embedded accelerometers with computer vision for recognizing food preparation activities},
  author={Stein, Sebastian and McKenna, Stephen J},
  booktitle={Proceedings of the 2013 ACM international joint conference on Pervasive and Ubiquitous Computing},
  pages={729--738},
  year={2013}
}

@inproceedings{alayrac2016unsupervised,
  title={Unsupervised learning from narrated instruction videos},
  author={Alayrac, Jean-Baptiste and Bojanowski, Piotr and Agrawal, Nishant and Sivic, Josef and Laptev, Ivan and Lacoste-Julien, Simon},
  booktitle={Proceedings of the IEEE Conference on Computer Vision and Pattern Recognition},
  pages={4575--4583},
  year={2016}
}

@inproceedings{kukleva2019unsupervised,
  title={Unsupervised learning of action classes with continuous temporal embedding},
  author={Kukleva, Anna and Kuehne, Hilde and Sener, Fadime and Gall, Jurgen},
  booktitle={Proceedings of the IEEE Conference on Computer Vision and Pattern Recognition},
  pages={12066--12074},
  year={2019}
}

@inproceedings{li2021action,
  title={Action shuffle alternating learning for unsupervised action segmentation},
  author={Li, Jun and Todorovic, Sinisa},
  booktitle={Proceedings of the IEEE Conference on Computer Vision and Pattern Recognition},
  pages={12628--12636},
  year={2021}
}

@inproceedings{caron2018deep,
  title={Deep clustering for unsupervised learning of visual features},
  author={Caron, Mathilde and Bojanowski, Piotr and Joulin, Armand and Douze, Matthijs},
  booktitle={Proceedings of the European Conference on Computer Vision},
  pages={132--149},
  year={2018}
}

@inproceedings{swetha2021unsupervised,
  title={Unsupervised discriminative embedding for sub-action learning in complex activities},
  author={Swetha, Sirnam and Kuehne, Hilde and Rawat, Yogesh S and Shah, Mubarak},
  booktitle={2021 IEEE International Conference on Image Processing},
  pages={2588--2592},
  year={2021},
  organization={IEEE}
}

@inproceedings{vidalmata2021joint,
  title={Joint visual-temporal embedding for unsupervised learning of actions in untrimmed sequences},
  author={VidalMata, Rosaura G and Scheirer, Walter J and Kukleva, Anna and Cox, David and Kuehne, Hilde},
  booktitle={Proceedings of the IEEE Winter Conference on Applications of Computer Vision},
  pages={1238--1247},
  year={2021}
}

@inproceedings{chao2018rethinking,
  title={Rethinking the faster r-cnn architecture for temporal action localization},
  author={Chao, Yu-Wei and Vijayanarasimhan, Sudheendra and Seybold, Bryan and Ross, David A and Deng, Jia and Sukthankar, Rahul},
  booktitle={Proceedings of the IEEE Conference on Computer Vision and Pattern Recognition},
  pages={1130--1139},
  year={2018}
}

@inproceedings{shou2017cdc,
  title={Cdc: Convolutional-de-convolutional networks for precise temporal action localization in untrimmed videos},
  author={Shou, Zheng and Chan, Jonathan and Zareian, Alireza and Miyazawa, Kazuyuki and Chang, Shih-Fu},
  booktitle={Proceedings of the IEEE Conference on Computer Vision and Pattern Recognition},
  pages={5734--5743},
  year={2017}
}

@article{li2020ms,
  title={Ms-tcn++: Multi-stage temporal convolutional network for action segmentation},
  author={Li, Shijie and Farha, Yazan Abu and Liu, Yun and Cheng, Ming-Ming and Gall, Juergen},
  journal={IEEE Transactions on Pattern Analysis and Machine Intelligence},
  volume={45},
  number={6},
  pages={6647--6658},
  year={2020},
  publisher={IEEE}
}

@inproceedings{farha2019ms,
  title={Ms-tcn: Multi-stage temporal convolutional network for action segmentation},
  author={Farha, Yazan Abu and Gall, Jurgen},
  booktitle={Proceedings of the IEEE Conference on Computer Vision and Pattern Recognition},
  pages={3575--3584},
  year={2019}
}

@inproceedings{lea2017temporal,
  title={Temporal convolutional networks for action segmentation and detection},
  author={Lea, Colin and Flynn, Michael D and Vidal, Rene and Reiter, Austin and Hager, Gregory D},
  booktitle={proceedings of the IEEE Conference on Computer Vision and Pattern Recognition},
  pages={156--165},
  year={2017}
}

@inproceedings{ahn2021refining,
  title={Refining action segmentation with hierarchical video representations},
  author={Ahn, Hyemin and Lee, Dongheui},
  booktitle={Proceedings of the IEEE International Conference on Computer Vision},
  pages={16302--16310},
  year={2021}
}

@inproceedings{ishikawa2021alleviating,
  title={Alleviating over-segmentation errors by detecting action boundaries},
  author={Ishikawa, Yuchi and Kasai, Seito and Aoki, Yoshimitsu and Kataoka, Hirokatsu},
  booktitle={Proceedings of the IEEE Winter Conference on Applications of Computer Vision},
  pages={2322--2331},
  year={2021}
}

@article{yi2021asformer,
  title={Asformer: Transformer for action segmentation},
  author={Yi, Fangqiu and Wen, Hongyu and Jiang, Tingting},
  journal={British Machine Vision Conference},
  year={2021}
}

@inproceedings{guan2021domain,
  title={Domain adaptive video segmentation via temporal consistency regularization},
  author={Guan, Dayan and Huang, Jiaxing and Xiao, Aoran and Lu, Shijian},
  booktitle={Proceedings of the IEEE International Conference on Computer Vision},
  pages={8053--8064},
  year={2021}
}

@inproceedings{behrmann2022unified,
  title={Unified fully and timestamp supervised temporal action segmentation via sequence to sequence translation},
  author={Behrmann, Nadine and Golestaneh, S Alireza and Kolter, Zico and Gall, J{\"u}rgen and Noroozi, Mehdi},
  booktitle={Proceedings of the European Conference on Computer Vision},
  pages={52--68},
  year={2022},
  organization={Springer}
}

@inproceedings{xu2024efficient,
  title={Efficient and Effective Weakly-Supervised Action Segmentation via Action-Transition-Aware Boundary Alignment},
  author={Xu, Angchi and Zheng, Wei-Shi},
  booktitle={Proceedings of the IEEE Conference on Computer Vision and Pattern Recognition},
  pages={18253--18262},
  year={2024}
}

@inproceedings{rahaman2019spectral,
  title={On the spectral bias of neural networks},
  author={Rahaman, Nasim and Baratin, Aristide and Arpit, Devansh and Draxler, Felix and Lin, Min and Hamprecht, Fred and Bengio, Yoshua and Courville, Aaron},
  booktitle={International Conference on Machine Learning},
  pages={5301--5310},
  year={2019},
  organization={PMLR}
}

@article{xu2018understanding,
  title={Understanding training and generalization in deep learning by fourier analysis},
  author={Xu, Zhiqin John},
  journal={arXiv preprint arXiv:1808.04295},
  year={2018}
}

@article{ronen2019convergence,
  title={The convergence rate of neural networks for learned functions of different frequencies},
  author={Ronen, Basri and Jacobs, David and Kasten, Yoni and Kritchman, Shira},
  journal={Advances in Neural Information Processing Systems},
  volume={32},
  year={2019}
}

@article{tancik2020fourier,
  title={Fourier features let networks learn high frequency functions in low dimensional domains},
  author={Tancik, Matthew and Srinivasan, Pratul and Mildenhall, Ben and Fridovich-Keil, Sara and Raghavan, Nithin and Singhal, Utkarsh and Ramamoorthi, Ravi and Barron, Jonathan and Ng, Ren},
  journal={Advances in Neural Information Processing Systems},
  volume={33},
  pages={7537--7547},
  year={2020}
}

@article{zagoruyko2017diracnets,
  title={Diracnets: Training very deep neural networks without skip-connections},
  author={Zagoruyko, Sergey and Komodakis, Nikos},
  journal={arXiv preprint arXiv:1706.00388},
  year={2017}
}

@inproceedings{ali2025joint,
  title={Joint self-supervised video alignment and action segmentation},
  author={Ali, Ali Shah and Mahmood, Syed Ahmed and Saeed, Mubin and Konin, Andrey and Zia, M Zeeshan and Tran, Quoc-Huy},
  booktitle={Proceedings of the IEEE/CVF International Conference on Computer Vision},
  pages={10807--10818},
  year={2025}
}

@inproceedings{hvq2025spurio,
    author    = {Federico Spurio and Emad Bahrami and Gianpiero Francesca and Juergen Gall},
    title     = {Hierarchical Vector Quantization for Unsupervised Action Segmentation},
    booktitle = {AAAI Conference on Artificial Intelligence (AAAI)},
    year      = {2025}
}

@inproceedings{bueno2025clot,
  title={Clot: Closed loop optimal transport for unsupervised action segmentation},
  author={Bueno-Benito, Elena and Dimiccoli, Mariella},
  booktitle={Proceedings of the IEEE/CVF International Conference on Computer Vision},
  pages={10719--10729},
  year={2025}
}

@article{li2026learning,
  title={Learning Probabilistic Embeddings for Unsupervised Action Segmentation},
  author={Li, Shuai and Vu, Duc Manh and Gall, Juergen},
  journal={arXiv preprint arXiv:2607.05263},
  year={2026}
}

@article{song2025unsupervised,
  title={Unsupervised Action Segmentation via Multi-scale Temporal-interaction Enhancement},
  author={Song, Zhiying and Chen, Kaixuan and Wang, Pengfei and Song, Mingli and Zheng, Nenggan},
  journal={IEEE Transactions on Circuits and Systems for Video Technology},
  year={2025},
  publisher={IEEE}
}

@article{zhao2025spike,
  title={Spike camera optical flow estimation based on continuous spike streams},
  author={Zhao, Rui and Xiong, Ruiqin and Wang, Dongkai and Xuan, Shiyu and Zhang, Jian and Fan, Xiaopeng and Huang, Tiejun},
  journal={IEEE Transactions on Pattern Analysis and Machine Intelligence},
  year={2025},
  publisher={IEEE}
}

@inproceedings{zhao2026featjnd,
  title={Just Noticeable Difference Modeling for Deep Visual Features},
  author={Zhao, Rui and Li, Wenrui and Zhu, Lin and Zheng, Yajing and Lin, Weisi},
  booktitle={International Conference on Machine Learning (ICML)},
  year={2026}
}

@article{ju2024deep,
  title={Deep learning methods for calibrated photometric stereo and beyond},
  author={Ju, Yakun and Lam, Kin-Man and Xie, Wuyuan and Zhou, Huiyu and Dong, Junyu and Shi, Boxin},
  journal={IEEE Transactions on Pattern Analysis and Machine Intelligence},
  volume={46},
  number={11},
  pages={7154--7172},
  year={2024},
  publisher={IEEE}
}

@article{ju2025revisiting,
  title={Revisiting one-stage deep uncalibrated photometric stereo via fourier embedding},
  author={Ju, Yakun and Shi, Boxin and Wen, Bihan and Lam, Kin-Man and Jiang, Xudong and Kot, Alex C},
  journal={IEEE Transactions on Pattern Analysis and Machine Intelligence},
  volume={47},
  number={8},
  pages={6185--6199},
  year={2025},
  publisher={IEEE}
}

@article{li2024object,
  title={Object segmentation-assisted inter prediction for versatile video coding},
  author={Li, Zhuoyuan and Yuan, Zikun and Li, Li and Liu, Dong and Tang, Xiaohu and Wu, Feng},
  journal={IEEE Transactions on Broadcasting},
  volume={70},
  number={4},
  pages={1236--1253},
  year={2024},
  publisher={IEEE}
}

@article{li2025ustc,
  title={USTC-TD: A test dataset and benchmark for image and video coding in 2020s},
  author={Li, Zhuoyuan and Liao, Junqi and Tang, Chuanbo and Zhang, Haotian and Li, Yuqi and Bian, Yifan and Sheng, Xihua and Feng, Xinmin and Li, Yao and Gao, Changsheng and others},
  journal={IEEE Transactions on Multimedia},
  year={2025},
  publisher={IEEE}
}

\end{document}